\documentclass{article}
\usepackage{spconf,amsmath,epsfig}
\usepackage{url}
\usepackage{subfig}
\usepackage{tabularx}
\usepackage{multirow}
\usepackage{booktabs}

\title{MULTI-SPECTRAL REMOTE SENSING IMAGE RETRIEVAL\\USING GEOSPATIAL FOUNDATION MODELS}
\fourauthors
  {Benedikt Blumenstiel\thanks{
  Co-funded by the European Union (Horizon Europe, Embed2Scale, 101131841).
  The corresponding author can be reached via email at benedikt.blumenstiel@ibm.com.
\\ © 2024 IEEE. Personal use of this material is permitted. Permission from IEEE must be obtained for all other uses, in any current or future media, including reprinting/republishing this material for advertising or promotional purposes, creating new collective works, for resale or redistribution to servers or lists, or reuse of any copyrighted component of this work in other works.
}}
  {IBM Research\\Europe}
  {Viktoria Moor}
  {IBM Research\\Europe}
  {Romeo Kienzler}
  {IBM Research\\Europe}
  {Thomas Brunschwiler}
  {IBM Research\\Europe}

\begin{document}
%
\maketitle

\begin{abstract}

Image retrieval enables an efficient search through vast amounts of satellite imagery and returns similar images to a query. 
Deep learning models can identify images across various semantic concepts without the need for annotations.
This work proposes to use Geospatial Foundation Models, like Prithvi, for remote sensing image retrieval with multiple benefits: i) the models encode multi-spectral satellite data and ii) generalize without further fine-tuning. 
We introduce two datasets to the retrieval task and observe a strong performance: Prithvi processes six bands and achieves a mean Average Precision of 97.62\% on BigEarthNet-43 and 44.51\% on ForestNet-12, outperforming other RGB-based models. Further, we evaluate three compression methods with binarized embeddings balancing retrieval speed and accuracy. They match the retrieval speed of much shorter hash codes while maintaining the same accuracy as floating-point embeddings but with a 32-fold compression.
The code is available at \url{https://github.com/IBM/remote-sensing-image-retrieval}.
\end{abstract}
\section{Introduction}
\label{sec:intro}

Remote sensing image retrieval has become increasingly essential in geospatial data analysis, with its potential applications extending across meteorology~\cite{meteorological2021}, economic assessment~\cite{economic2014}, and ecological analysis~\cite{ecology2011}. 
In recent years, machine learning enabled a shift from traditional metadata-based retrieval methods to content-based image retrieval (CBIR)~\cite{Survey2021}. CBIR focuses on the intrinsic features within a query image. This enables the retrieval of potentially any semantic concept without requiring pre-defined annotations in metadata. 

Central to CBIR are retrieval speed and accuracy. While accuracy ensures the relevance of the retrieved images to the query, speed is crucial for efficient processing in large-scale databases. To balance these metrics, compression techniques, such as hash functions, are used to reduce memory requirements and increase retrieval speed.
Promising works have combined pre-trained vision models with deep hash networks~\cite{AsymmetricHash2022,Milan2021,CrossModalIR2022,MetaHash2022}. Each image is represented by the model embedding, which is compressed into a hash code. However, these models only process RGB data, overlooking the potential of multi-spectral information in satellite imagery.

Geospatial Foundation Models (GeoFM), for instance Prithvi~\cite{Prithvi2023} and SatMAE~\cite{SatMAE2022}, open new possibilities because they are pre-trained on a vast amount of multi-spectral data. The models have been applied to various tasks, including flood and wildfire segmentation~\cite{Prithvi2023}. GeoFMs can utilize multi-spectral data for CBIR without requiring fine-tuning.

Our contributions are threefold: i)~We showcase the application of GeoFMs for remote sensing image retrieval,
ii)~we introduce baselines for two multi-spectral datasets to benchmark multi-spectral remote sensing image retrieval, and 
iii)~we conduct a detailed analysis of retrieval performance between vector-based and binary hash-based approaches, focusing on the balance between speed and accuracy.

\section{Related work}
\label{sec:relatedwork}

Image retrieval in remote sensing has evolved significantly, shifting focus from traditional metadata-based methods to more advanced techniques that use computer vision to analyze the image content~\cite{Survey2021}. 
Among these developments, deep hash networks have played a central role~\cite{AsymmetricHash2022,MetaHash2022,MultiScaleHash2023,AdversarialHash2020}. These models compress images into hash codes which are stored in a database and used for calculating distances. Researchers explored various learning techniques, such as contrastive learning~\cite{MetaHash2022} and metric learning~\cite{MetricLeanring2018}, with the aim of minimizing the need for extensive training data.

Some approaches use the embeddings of existing computer vision models and combine them with smaller hash models~\cite{AsymmetricHash2022,Milan2021,MetricLeanring2018}. Because these models are trained on RGB data, they cannot take advantage of all multi-spectral data from satellites. Furthermore, the fine-tuning of hash networks on specific datasets limits their transferability to different semantic concepts, potentially reducing accuracy in varying contexts beyond their initial training datasets.

Despite the progress in various CBIR techniques, most evaluations are primarily focused on RGB datasets, i.e., UCM~\cite{UCM2010} and AID~\cite{AID2017} being the \textit{de-facto} standard evaluation~\cite{AsymmetricHash2022,Milan2021,MetaHash2022,MultiScaleHash2023,AdversarialHash2020,MetricLeanring2018,ContrastiveLHash2023,DeepHashing2018}.
An exception is works on cross-modal image retrieval~\cite{Survey2021}. E.g., one study uses a subset of BigEarthNet~\cite{BigEarthNet2019}, which includes multi-spectral data~\cite{CrossModalIR2022}.

GeoFMs, such as SatMAE, Prithvi, and Presto~\cite{Presto2023}, have the ability to process multi-spectral data. Their pre-training on diverse satellite images overcomes the limitations of general models and benefits earth observation tasks~\cite{Prithvi2023}.

In conclusion, while existing approaches have advanced the field, they are focused on RGB datasets and depend on annotated training data. These limitations emphasize the need for new approaches utilizing GeoFMs and multi-spectral data. 

\section{Approach}
\label{sec:approach}

The CBIR task requires identifying and retrieving images from a large database based on their visual and semantic similarity to a given query image. This process requires an efficient mechanism to compare and rank the database images in terms of their similarity to the query.

\begin{figure}[tbh]
  \centering
  \begin{minipage}[b]{\linewidth}
  \includegraphics[width=\linewidth]{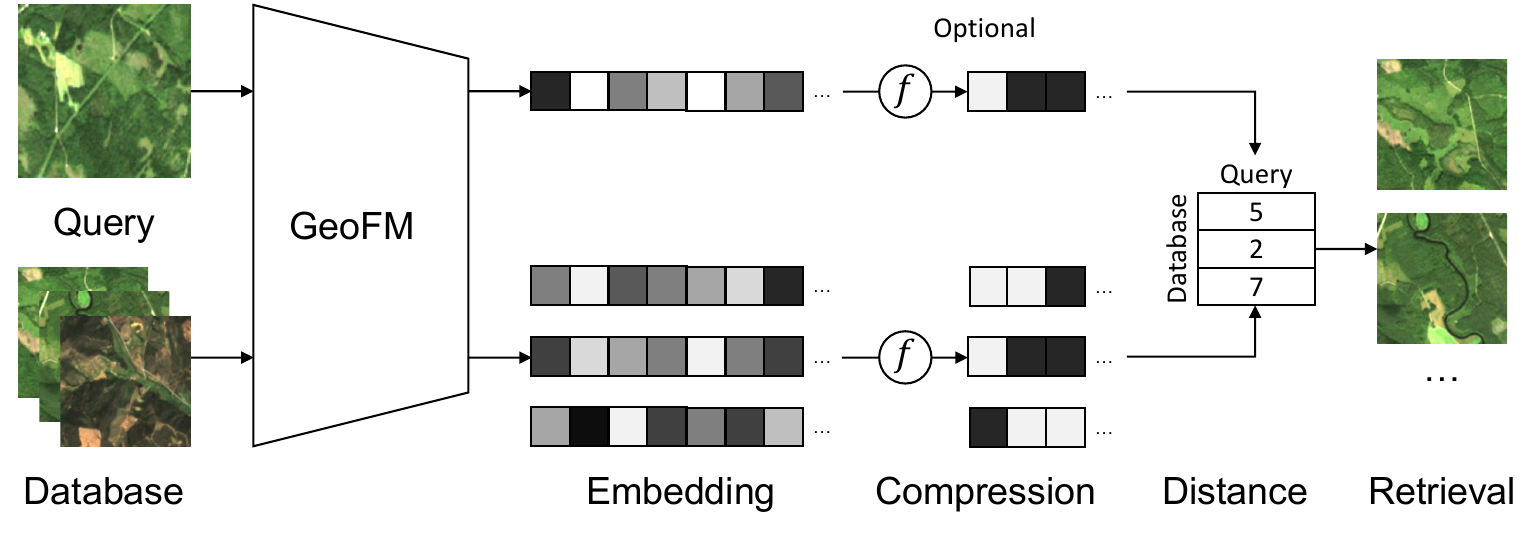}
  \end{minipage}
\caption{GeoFM embeddings enable simple but accurate CBIR. Optionally, the embeddings are compressed into smaller binary vectors. For each query image, similar images from the database are returned and sorted based on a distance function.}
\label{fig:res}
\end{figure}

In our method, as illustrated in Figure \ref{fig:res}, the first step involves processing the query images through a GeoFM. The model generates embedding vectors that comprehensively represent latent features of the images. The query embedding is compared with the pre-computed embeddings stored in a vector database. The similarity between query vectors and database vectors is computed using common distance functions like Hamming, Jaccard, Euclidean or Cosine distance.
This approach takes advantage of the robust pre-training of the underlying GeoFM, which improves the accuracy and generalizability of the retrieval results. Optionally, the vector embeddings are compressed into shorter binary vectors using existing methods like hash functions. This reduces memory usage and inference time but can lead to information loss.

The simplest compression is the binarization of the embedding vector through a sign function. This reduces the memory usage by a factor of 32. For further compression, we propose a trivial hash function as a simple baseline: The embeddings are split into an equally sized number of dimensions and averaged to reduce the vector to the hash length. We then apply the sign function to generate binary hash codes. This approach assumes an equal distribution of information across the dimensions and around zero within each dimension.

\newcolumntype{C}{>{\centering\arraybackslash}X}
\begin{table*}[tbh]
\begin{tabularx}{\linewidth}{llCCCCc}
\toprule
Model & Method & BigEarthNet-43 & BigEarthNet-19 & ForestNet-12 & ForestNet-4 & Mean \\
\midrule
\multirow[c]{3}{*}{Prithvi-100M} & Embedding & \textbf{97.62} & \textbf{97.98} & \textbf{44.51} & \textbf{60.76} & \textbf{75.22} \\
 & Binary emb. & \underline{97.44} & \underline{97.83} & \underline{43.28} & \underline{59.85} & \underline{74.6}\phantom{0} \\
 & 64-bit hash & 92.58 & 93.44 & 41.49 & 55.93 & 70.86 \\
 \hline
\multirow[c]{3}{*}{SatMAE-ViT-B} & Embedding & 94.78 & 95.59 & 37.61 & 52.94 & 70.23 \\
 & Binary emb. & 89.39 & 90.4\phantom{0} & 36.49 & 53.04 & 67.33 \\
 & 64-bit hash & 79.35 & 80.99 & 30.79 & 47.19 & 59.58 \\
\hline
\multirow[c]{3}{*}{Prithvi-100M-RGB} & Embedding & 92.15 & 93.17 & 38.65 & 53.85 & 69.46 \\
 & Binary emb. & 91.38 & 92.43 & 38.11 & 53.31 & 68.81 \\
 & 64-bit hash & 82.60 & 84.45 & 32.58 & 48.20 & 61.96 \\
\hline
 \multirow[c]{3}{*}{ViT-B/16-RGB}& Embedding & 89.31 & 90.21 & 38.92 & 56.49 & 68.73 \\
  & Binary emb. & 88.71 & 89.7\phantom{0} & 39.19 & 57.01 & 68.65 \\
  & 64-bit hash & 79.01 & 81.54 & 33.60 & 49.63 & 60.95 \\

\bottomrule
\end{tabularx}

\caption{mAP@20 results for all models and datasets. We highlight the best-performing method in bold and underline the second-best one. The 64-bit hash uses the described trivial hash approach by averaging twelve embedding values for each bit.}
\label{tab:results}
\vspace{-3mm}
\end{table*}

\section{Experimental setup}
\label{sec:experiments}

Following literature~\cite{AsymmetricHash2022,Milan2021,MetaHash2022,MultiScaleHash2023,AdversarialHash2020,MetricLeanring2018,ContrastiveLHash2023,DeepHashing2018}, we evaluate our experiments with mean Average Precision (mAP) based on the top 20 retrieved images. For multi-label datasets, any overlap within the labels is counted as a positive match~\cite{CrossModalIR2022,MultiLabel2020}.
Each validation split serves as test queries, and the test split is used as the database. We use different splits to avoid a geographical overlap between the queries and the database\footnote{
Other works iterate over the test split as query images, using the remaining test images as the database~\cite{Milan2021,MetricLeanring2018}. This is appropriate for aerial datasets but not for satellite datasets. Some splits are geographically grouped to avoid similar regional patterns benefiting the model performance. 
We also find some implementations using train images as part of the queries or database~\cite{CrossModalIR2022,MetaHash2022,ContrastiveLHash2023,DeepHashing2018}. Using the same images during training and testing can lead to data leakage, which benefits over-fitted models. 

Therefore, we select two different splits for the queries and databases. Note that we do not train any model since we use pre-trained GeoFMs. However, we avoid using the train splits to facilitate a fair comparison with potential future works that use trained models. For training, we recommend further splitting the train images and not using the official validation splits.}.
Further, we use the L1 norm as a distance function in our experiments, which is equal to the hamming distance for binary values. We also tested the L2 norm and observed minimal differences within $\pm$ 1 pp. mAP.

We use Milvus~\cite{Milvus2021} for our speed experiments. Milvus is a production-ready vector database and includes a search functionality based on binary and floating-point vectors. The L2 norm is used for float vectors, as L1 is only available for binary vectors. Milvus indexes\footnote{See \url{https://milvus.io/docs/index.md} for details. We used the indexes INV\_FLOAT and BIN\_IVF\_FLAT with the default values of 128 clusters and 8 top clusters.}
include a cluster-based approach: First, the query is compared to the cluster centers, followed by a comparison with the images in the top clusters. This drastically reduces the retrieval time.

\subsection{Models}
\label{sec:models}

We evaluate Prithvi-100M and SatMAE-ViT-B as the underlying GeoFMs.
Prithvi-100M is a Vision Transformer (ViT)~\cite{ViT2020} with 100M parameters and a 224x224 pixel input. Prithvi processes six bands, unlike the three RGB channels of the vanilla ViT. The pre-training includes a subset of the Harmonized Landsat-Sentinel (HLS) dataset consisting of images from the Landsat~8 and Sentinel-2 satellites at 30~meter resolution. We also report the results of Prithvi only processing the RGB channels for comparison (Prithvi-100M-RGB). 
The infrared channels are set to the channel mean which is mathematically similar to dropping these channels.
SatMAE is a ViT-B/16 model with a 96x96 pixel input. The model is trained on ten Sentinel-2 bands at 10~meter resolution. For Landsat images, we fill the missing bands with the channel mean similar to Prithvi-100M-RGB.
For comparison, we include the vanilla ViT-B/16 model (ViT-B/16-RGB)~\cite{ViT2020}, which is pre-trained on ImageNet-21K.

We run the experiments with the model embeddings of size 768, binary embeddings, and trivial hash codes of length 64.
We varied the hash length in additional experiments and find that 64 bit is a good compromise between compression and accuracy, as the performance drops significantly with a smaller hash length.
We also tested Locality-Sensitive Hashing (LSH)~\cite{LSH2017} but did not observed a overall improved performance compared to the trivial approach. 

\subsection{Datasets}
\label{sec:datasets}

Reviewing remote sensing datasets revealed a limited availability of multi-spectral multi-class datasets with an sufficient image size \cite{Survey2021,Earthnets2022,GeoBench2023}. BigEarthNet~\cite{BigEarthNet2019} and ForestNet~\cite{ForestNet2020} fulfill these requirements.

BigEarthNet consists of Sentinel-1 and Sentinel-2 images with 120x120 pixels and includes two sets of multi-label annotations with 19 resp. 43 classes. The classes cover different land-use types such as \textit{mixed forest}, \textit{water bodies}, or \textit{airports}. We are using the six resp. ten Sentinel-2 channels supported by Prithvi and SatMAE. 

ForestNet includes Landsat 8 imagery from forest loss events with a 332x332 image size. The images are annotated with twelve classes and four super-classes. The classes indicate types of deforestation, such as \textit{timber plantation} or \textit{small-scale agriculture}. 
ForestNet uses composite images created by averaging up to five cloud-free images. 

Note that the data of both datasets differs from the pre-training settings. The images are bi-linear re-scaled to match the model input size, which leads to different spatial resolutions than during pre-training. Furthermore, SatMAE was not trained on Landsat data and Prithvi-100M is pre-trained only on US data while BigEarthNet and ForestNet cover Europe and Indonesia, resp.~\cite{Prithvi2023,BigEarthNet2019,ForestNet2020}.

\begin{figure}[b]
 \vspace{-3mm}
  \centering
  \subfloat[][Embedding]{
   \includegraphics[width=0.225\linewidth]{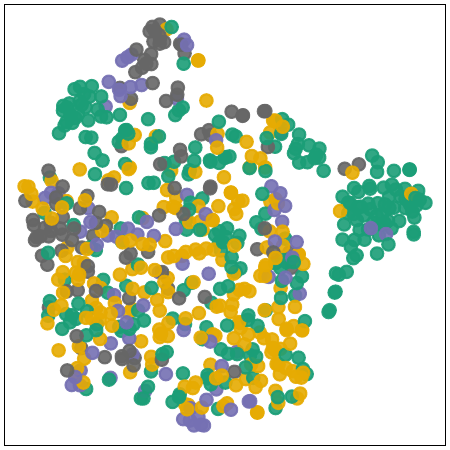}
  }
  \subfloat[][Binary emb.]{
   \includegraphics[width=0.225\linewidth]{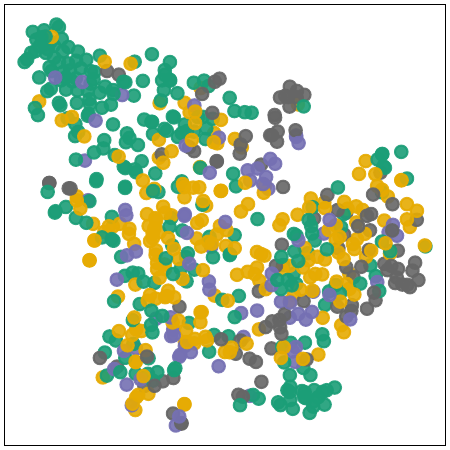}
  }
  \subfloat[][64-bit hash]{
   \includegraphics[width=0.225\linewidth]{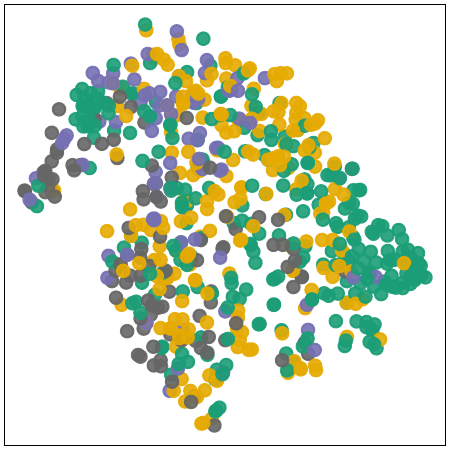}
  }
  \subfloat[][32-bit hash]{
   \includegraphics[width=0.225\linewidth]{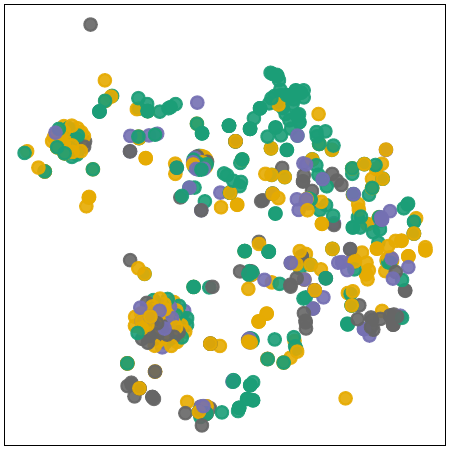}
  }

\caption{t-SNE plots of the ForestNet-4 test set with colored classes comparing Pritvi-100M embeddings.}
\label{fig:tsne}
\end{figure}

\begin{figure*}[tb]
  \centering
  \subfloat[][BigEarthNet-43]{
   \includegraphics[width=0.99\linewidth]{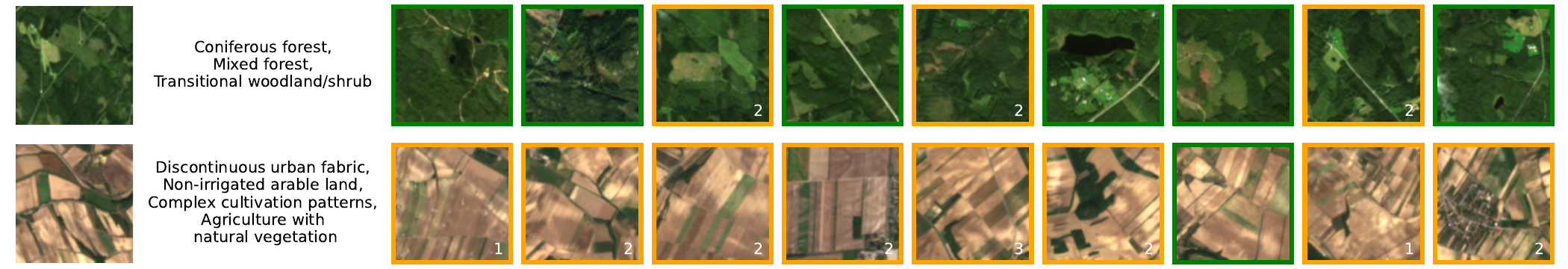}
  } \\
  \vspace{-3mm}
  \subfloat[][ForestNet-12]{
   \includegraphics[width=0.99\linewidth]{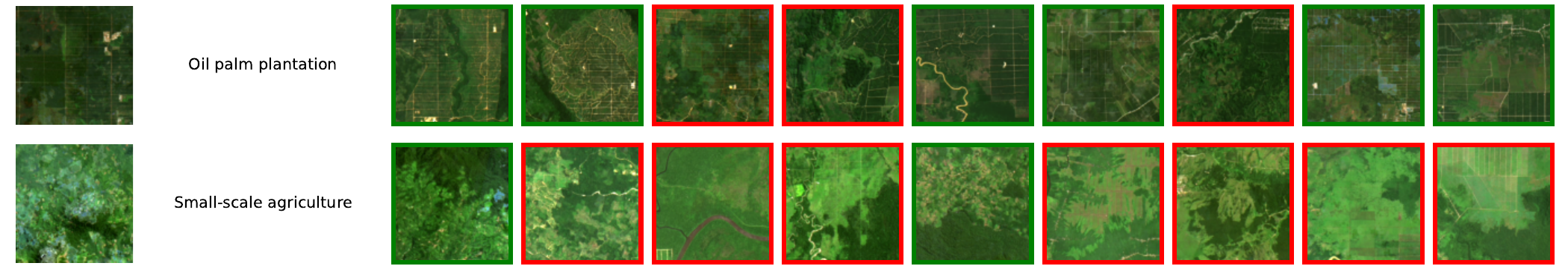}
  }
\vspace{-2mm}
\caption{Examples from two datasets with query images (left), their labels, and retrieved images (right) using Prithvi-100M and the 64-bit trivial hash. Images with green frames indicate positive matches, while those with red frames have different labels. Orange shows partial correct matches, where the number represents the number of label matches within the multi-labels.}
\label{fig:examples}
\vspace{-3mm}
\end{figure*}

\section{Results}
\label{sec:results}

Prithvi-100M outperforms all other models in every method by over 4.9 pp. on average as shown in Table~\ref{tab:results}. 
The transition from float embeddings to hash-based approaches results in a notable drop in model performance. E.g., Prithvi-100M with a 64-bit trivial hash has on average an 4.36 pp. lower mAP. The difference is larger for other models: SatMAE and vanilla ViT have a average difference of 10.65 and 7.78 pp. resp. This drop highlights the challenge of maintaining accuracy while simplifying the data representations. Except for SatMAE, we observe that the performance loss is mainly influenced by the hash length and not by binarization. The binary embeddings perform nearly as well as the floating-point embeddings, with an average difference of only 0.62 pp. mAP for Prithvi-100M.

While SatMAE outperforms the RGB-only models on BigEarthNet, it performs worse on ForestNet. The missing channels and different data source might affect the performance. 
Overall, the comparison highlights the information gain when using multi-spectral data with GeoFMs that are trained on satellite data.



We visually compare the ForestNet-4 embeddings and hash functions with t-SNE plots in Figure~\ref{fig:tsne}. The latent space is well distributed for the float and binarized embeddings as well as the 64-bit trivial hash. The 32-bit trivial hash has clusters which do not discriminate the semantic classes and explain the much lower performance of shorter hash codes.

Figure~\ref{fig:examples} shows examples of retrieved images based on the trivial hashes. Retrieved images from BigEarthNet-43 have a large share of images with partial label matches, which is also represented by a 92.58 mAP. Those images are often similar in their overall color and appearance. However, some classes are rarely matched, e.g., specific infrastructure classes like \textit{construction sites} or \textit{airports}. This is also observable in ForestNet-12, which has more fine-grained classes that are harder to differentiate. Still, our approach returns some matching examples for most queries from this dataset. 

Our analysis demonstrated the effect of compression on the accuracy. Next, we discuss the effect on retrieval speed.
The overall retrieval speed depends on three factors: i) model inference, ii) similarity search, and iii) data loading. The first step mainly depends on the model used for the encoding. We observed an inference time of ~100 ms for Prithvi on a NVIDIA V100 GPU. Using smaller GeoFMs can reduce the processing time. The last step depends on the specific hardware and is not affected by the retrieval approach. Therefore, we focus our speed analysis on the actual retrieval computation. We used a VM with 12~cores and 24~Gb of memory on an AMD EPYC 7452 processor for our experiments and provide the results in Table~\ref{tab:speed}.

\begin{table}[h]
\centering
\begin{tabularx}{\linewidth}{lCCCC}
\toprule
 & & \multicolumn{3}{c}{Images in database}\\
Data type & Length & 10K & 50K & 100K \\
\midrule
Binary & 64 & 16 ms & 16 ms & 17 ms \\
Binary & 768 & 16 ms & 16 ms & 17 ms \\
Float & 768 & 21 ms & 32 ms & 33 ms \\
\bottomrule
\end{tabularx}

\caption{Experimental retrieval speeds with different vector types for a varying number of images in the database.}
\label{tab:speed}
\vspace{-1.5mm}
\end{table}

The retrieval speed for binary vectors is almost not influenced by the database size or the vector length. Floating-point embeddings have a retrieval time of up to 33~ms. This is two times longer than binary embeddings with 17~ms. The results demonstrate the efficient implementation of retrieval algorithms in vector databases like Milvus by using cluster centers. Therefore, the hash length is mainly constrained by memory usage rather than the retrieval speed.

\section{Conclusion}
\label{sec:conclusion}
This work demonstrates the applicability of GeoFMs for image retrieval in remote sensing. Due to the learned representations through the pre-training, Prithvi-100M encodes multiple semantics and does not require further fine-tuning. We introduce two multi-spectral datasets to the retrieval task and provide strong baselines, enabling a more holistic evaluation for remote sensing in the future. 
Accordingly, we evaluated two compression methods and binary embeddings have shown the best trade-off between accuracy and retrieval speed.
The approach can be easily implemented in various applications with any existing GeoFM and combined with more advanced compression methods to improve performance.

\vfill
\pagebreak

\bibliographystyle{IEEEbib}
\bibliography{refs}

\end{document}